\documentclass{article}

\usepackage{times}
\usepackage{epsfig, graphics, subfigure}
\usepackage{subfig}
\usepackage{latexsym}
\usepackage{amsmath,amssymb,amsthm,amsfonts}
\usepackage[numbers]{natbib}
\usepackage{algorithm,algorithmic}

\title{On Learning Discrete Graphical Models Using Greedy Methods}

\author{Ali Jalali\\
ECE, University of Texas at Austin\\
\texttt{alij@mail.utexas.edu}\\
\and
Chris Johnson\\
CS, University of Texas at Austin\\
\texttt{cjohnson@cs.utexas.edu}\\
\and
Pradeep Ravikumar\\
CS, University of Texas at Austin\\
\texttt{pradeepr@cs.utexas.edu}\\}

\def\tparam{\theta^{*}}
\def\paramset{\real^\pdim}
\def\set{\mathbb{S}}

\newcommand{\inprod}[2]{\langle {#1},  {#2} \rangle}

\def\sgn{\text{Sign}}

\newtheorem{lemma}{Lemma}
\newtheorem{theorem}{Theorem}

\def\Data{D}
\def\numobs{n}

\def\L{\mathcal{L}}
\def\real{\mathbb{R}}

\def\Nr{\N_\svert}
\def\Loss{\mathcal{L}}

\def\param{\theta}
\def\paramHat{\widehat{\param}}
\def\paramStar{\param^*}
\def\supp{S}
\def\suppHat{\widehat{\supp}}
\def\suppStar{\supp^*}
\def\suppUnion{\supp}
\def\EpsilonStop{\epsilon_{\mathcal{S}}}
\def\LipshitzUpper{\kappa_{u}}
\def\LipshitzLower{\kappa_{l}}
\def\delpar{\Delta}
\def\delparHat{\widehat{\Delta}}
\def\grad{\nabla}
\def\noiseLevel{\lambda_{n}}
\def\union{\cup}
\def\ConstantNoise{c_{n}}

\newcommand\myparagraph[1]{\noindent {\it #1.}}
\newcommand{\tr}[2]{\left<#1,#2\right>}
\newcommand{\comment}[1]{}
\def\vertex{V}
\def\edge{E}
\def\svert{\ensuremath{r}}
\def\defn{:=}

\def\kStar{k^*}

\def\paramHatk{\widehat{\theta}^{(k)}}
\def\suppHatk{\widehat{S}^{(k)}}
\def\paramHatkm{\widehat{\theta}^{(k-1)}}
\def\suppHatkm{\widehat{S}^{(k-1)}}

\def\Data{\mathcal{D}}

\def\L{\mathcal{L}}

\def\real{\mathbb{R}}

\def\pdim{p}

\def\numobs{n}
\def\Data{D}

\def\vertex{V}
\def\edge{E}
\def\statenum{m}
\def\svert{\ensuremath{r}}
\def\defn{:=}

\def\graph{G}
\def\degmax{d}
\def\order{\mathcal{O}}
\def\mprob{\mathbb{P}}
\def\defn{:=}

\def\LipshitzUpper{\kappa_u}
\def\LipshitzLower{\kappa_l}

\def\Nr{\mathcal{N}_\svert}

\begin{document}
\maketitle
\begin{abstract}
In this paper, we address the problem of learning the structure of a pairwise graphical model from samples in a high-dimensional setting. Our first main result studies the sparsistency, or consistency in sparsity pattern recovery, properties of a forward-backward greedy algorithm as applied to general statistical models. As a special case, we then apply this algorithm to learn the structure of a discrete graphical model via neighborhood estimation. As a corollary of our general result, we derive sufficient conditions on the number of samples $n$, the maximum node-degree $d$ and the problem size $p$, as well as other conditions on the model parameters, so that the algorithm recovers all the edges with high probability. Our result guarantees graph selection for samples scaling as $n = \Omega(d^2 \, \log(p))$, in contrast to existing convex-optimization based algorithms that require a sample complexity of $\Omega(d^3\log(p))$. Further, the greedy algorithm only requires a restricted strong convexity condition which is typically milder than irrepresentability assumptions. We corroborate these results using numerical simulations at the end. 
\end{abstract}

\vspace{-0.6cm}
\section{Introduction} 

\vspace{-0.2cm}
Undirected graphical models, also known as Markov random fields, are used in a variety of domains, including statistical physics, natural language processing and image analysis among others. In this paper
we are concerned with the task of estimating the graph structure $G$ of a Markov random field (MRF) over a discrete random vector $X = (X_1, X_2,\ldots,X_p)$, given $\numobs$ independent and identically distributed samples $\{ x^{(1)}, x^{(2)}, \ldots, x^{(\numobs)} \}$. This underlying graph structure encodes conditional independence assumptions among subsets of the variables, and thus plays an important role in a broad range of applications of MRFs.

\myparagraph{Existing approaches: Neighborhood Estimation, Greedy Local Search} Methods for estimating such graph structure include those based on constraint and hypothesis testing~\citep{spirtes:00}, and those that estimate restricted classes of graph structures such as trees~\citep{chowliu:68}, polytrees~\cite{dasgupta:99}, and hypertrees~\citep{srebro:03}. A recent class of successful approaches for graphical model structure learning are based on estimating the local neighborhood of each node. One subclass of these for the special case of bounded degree graphs involve the use of exhaustive search so that their computational complexity grows at least as quickly as $\order(\pdim^{\degmax})$, where $\degmax$ is the maximum neighborhood size in the graphical model~\cite{AbbKolNg06, Bresler08, Csiszar:06}. Another subclass use convex programs to learn the neighborhood structure: for instance \citep{RWLIsing,Meinshausen:06,LeeKoller07} estimate the neighborhood set for each vertex $\svert \in V$ by optimizing its $\ell_1$-regularized conditional likelihood; \citep{JRVS11,Dahinden10} use $\ell_1/\ell_2$-regularized conditional likelihood. Even these methods, however need to solve regularized convex programs with typically polynomial computational cost of $O(p^4)$ or $O(p^6)$, are still expensive for large problems. Another popular class of approaches are based on using a score metric and searching for the best scoring structure from a candidate set of graph structures. Exact search is typically NP-hard~\cite{chickering:95}; indeed for general discrete MRFs, not only is the search space intractably large, but calculation of typical score metrics itself is computationally intractable since they involve computing the partition function associated with the Markov random field~\cite{Welsh93}. Such methods thus have to use approximations and search heuristics for tractable computation. 
\emph{Question:} Can one use local procedures that are as inexpensive as the heuristic greedy approaches, and yet come with the strong statistical guarantees of the regularized convex program based approaches?

\myparagraph{High-dimensional Estimation; Greedy Methods}
There has been an increasing focus in recent years on high-dimensional statistical models where the number of parameters $\pdim$ is comparable to or even larger than the number of observations $n$. 
It is now well understood that consistent estimation is possible even under such high-dimensional scaling if some low-dimensional structure is imposed on the model space. Of relevance to graphical model structure learning is the structure of sparsity, where a sparse set of non-zero parameters entail a sparse set of edges. A surge of recent work~\cite{CandesTao06,Donoho:Elad:03} has shown that $\ell_1$-regularization for learning such sparse models can lead to practical algorithms with strong theoretical guarantees. A line of recent work (cf. paragraph above) has thus leveraged this sparsity inducing nature of $\ell_1$-regularization,
to propose and analyze convex programs based on regularized log-likelihood functions. A related line of recent work on learning sparse models has focused on ``stagewise'' greedy algorithms.
These perform simple forward steps (adding parameters greedily), and possibly also backward steps (removing parameters greedily), and yet provide strong statistical guarantees for the estimate after a finite number of greedy steps. The forward greedy variant which performs just the forward step has appeared in various guises in multiple communities: in machine learning as boosting~\cite{Friedman00}, in function approximation~\cite{Temlyakov08}, and in signal processing as basis pursuit~\cite{Chen98}. In the context of statistical model estimation, \citet{Zhang09} analyzed the forward greedy algorithm for the case of sparse linear regression; and showed that the forward greedy algorithm is sparsistent (consistent for model selection recovery) under the same ``irrepresentable'' condition as that required for ``sparsistency'' of the Lasso. \citet{Zhang08} analyzes a more general greedy algorithm for sparse linear regression that performs forward and backward steps, and showed that it is sparsistent under a weaker restricted eigenvalue condition. 
Here we ask the question: \emph{Can we provide an analysis of a general forward backward algorithm for parameter estimation in general statistical models?} Specifically, we need to extend the sparsistency analysis of \cite{Zhang09} to general non-linear models, which requires a subtler analysis due to the circular requirement of requiring to control the third order terms in the Taylor series expansion of the log-likelihood, that in turn requires the estimate to be well-behaved. Such extensions in the case of $\ell_1$-regularization occur for instance in \cite{RWLIsing, Geer08, Bach10a}.

\myparagraph{Our Contributions}
In this paper, we address both questions above. In the first part, we analyze the forward backward greedy algorithm~\cite{Zhang09} for general statistical models. We note that even though we consider the 
general statistical model case, our analysis is much simpler and accessible than \cite{Zhang09}, and would be of use even to a reader interested in just the linear model case of \citet{Zhang09}. In the second part, we use this to show that when combined with neighborhood estimation, the forward backward variant applied to local conditional log-likelihoods provides a simple computationally tractable method that adds and deletes edges, but comes with strong \emph{sparsistency} guarantees. We reiterate that the our first result on the sparsistency of the forward backward greedy algorithm for general objectives is of independent interest even outside the context of graphical models. As we show, the greedy method is better than the $\ell_1$-regularized counterpart in \cite{RWLIsing} theoretically, as well as experimentally. The sufficient condition on the parameters imposed by the greedy algorithm is a restricted strong convexity condition~\cite{NRWY10}, which is weaker than the irrepresentable condition required by \cite{RWLIsing}. Further,
the number of samples required for sparsistent graph recovery scales as $O(d^2 \log p)$, where $d$ is the maximum node degree, in contrast to $O(d^3 \log p)$ for the $\ell_1$-regularized counterpart. We corroborate this in our simulations, where we find that the greedy algorithm requires fewer observations than \cite{RWLIsing} for sparsistent graph recovery.

\vspace{-0.2cm}
\section{Review, Setup and Notation}

\vspace{-0.25cm}
\subsection{Markov Random Fields} 

\vspace{-0.1cm}
Let $X = (X_1,\hdots,X_p)$ be a random vector, each variable $X_i$ taking values in a discrete set $\mathcal{X}$ of cardinality $m$. Let $G = (V,E)$ denote a graph with $p$ nodes,  corresponding to the $p$ variables $\{X_1,\hdots,X_p\}$. A pairwise Markov random field over $X=(X_1, \ldots, X_\pdim)$ is then specified by nodewise and pairwise functions
$\theta_{\svert} : \mathcal{X} \mapsto \real$ for all $\svert \in V$, and $\theta_{\svert t} : \mathcal{X} \times \mathcal{X} \mapsto \real$ for all $(\svert,t) \in E$:
\small\begin{equation}
\label{EqnGenMRFPairwise}
\mathbb{P}(x)  \propto \exp \big \{\sum_{\svert \in \vertex} \theta_{\svert}(x_\svert)  + \sum_{(\svert,t) \in \edge} \theta_{\svert t}(x_\svert,x_t) \big \}.
\end{equation}\normalsize
In this paper, we largely focus on the case where the variables are binary with $\mathcal{X} = \{-1,+1\}$, where we can rewrite \eqref{EqnGenMRFPairwise} to the Ising model form~\cite{Ising25} for some set of parameters $\{\theta_{\svert}\}$ and $\{\theta_{\svert t}\}$ as
\small\begin{equation}\label{EqnIsing}
 \mathbb{P}(x)  \propto  \exp \big \{\sum_{\svert \in \vertex} \theta_{\svert} x_\svert + \sum_{(\svert,t) \in \edge} \theta_{\svert t} x_\svert x_t \big \}.
\end{equation}\normalsize

%
%
%

\subsection{Graphical Model Selection}
Let $\Data \defn \{x^{(1)}, \ldots, x^{(\numobs)} \}$ denote the set of $\numobs$ samples, where each
$\pdim$-dimensional vector  $x^{(i)} \in \{1,\ldots,\statenum\}^\pdim$ is drawn i.i.d. from a distribution $\mprob_{\theta^*}$ of the form \eqref{EqnGenMRFPairwise}, for parameters $\theta^*$ and graph $\graph = (\vertex, \edge^*)$ over the $\pdim$ variables. Note that the true edge set $\edge^*$ can also be expressed as a function of the parameters as
\begin{align}
\label{EqnEdge}
\edge^* = \{(\svert,t) \in V \times V :\, \theta^*_{st} \neq 0 \}.
\end{align}
The \emph{graphical model selection} task consists of inferring this edge set $\edge^*$ from the samples $\Data$. The goal is to construct an estimator $\hat{E}_n$ for which $\mathbb{P}[\hat{E}_n = E^*] \rightarrow 1$ 
as $n \rightarrow \infty$. Denote by $\mathcal{N}^*(\svert)$ the set of neighbors of a vertex $\svert \in V$, so that $\mathcal{N}^*(\svert) = \{t : (\svert,t) \in E^*\}$. Then the graphical model selection problem is equivalent to that of estimating the neighborhoods $\hat{\mathcal{N}}_n(\svert) \subset \vertex$, so that $\mathbb{P}[\hat{\mathcal{N}}_n(\svert) = \mathcal{N}^*(\svert); \forall \svert \in \vertex]\rightarrow 1$ as $n \rightarrow \infty$. 

For any pair of random variables $X_\svert$ and $X_t$, the parameter $\theta_{\svert t}$ fully characterizes whether there is an edge between them, and can be estimated via its conditional likelihood. In particular, defining $\Theta_\svert \defn (\theta_{\svert 1},\ldots,\theta_{\svert p})$, our goal is to use the conditional likelihood of $X_\svert$ conditioned on $X_{\vertex\backslash\svert}$ to estimate 
$\Theta_\svert$ and hence its neighborhood $\mathcal{N}(\svert)$. This conditional distribution of $X_\svert$ conditioned on $X_{\vertex\backslash\svert}$ generated by \eqref{EqnIsing} is given by the logistic model
\small\begin{align*}
\mathbb{P}\left(X_\svert=x_\svert\Big|X_{\vertex\backslash\svert}=x_{\vertex\backslash\svert}\right) = \frac{\exp(\theta_\svert x_\svert + \sum_{t \in \vertex\backslash\svert} \theta_{\svert t}x_\svert x_t)}{1 + \exp(\theta_\svert + \sum_{\svert\in \vertex\backslash\svert} \theta_{\svert t} x_\svert)}.
\end{align*}\normalsize
Given the $n$ samples $\Data$, the corresponding conditional log-likelihood is given by
\small\begin{align}
\label{eq:Loss-Fn}
\Loss(\Theta_\svert;\Data) = \frac{1}{\numobs} \sum_{i=1}^\numobs \left\{ \log\!\left(\!1\! + \exp\left(\theta_\svert x^{(i)}\! +\!\!\!\! \sum_{t \in \vertex\backslash\svert} \theta_{\svert t} x^{(i)}_\svert x^{(i)}_t\right)\!\!\right)\!\!-\! \theta_\svert x^{(i)}_\svert\! -\!\!\!\! \sum_{t \in \vertex\backslash\svert} \theta_{\svert t} x^{(i)}_\svert x^{(i)}_t\right\}.
\end{align}\normalsize

In Section~\ref{SecAlgo}, we study a greedy algorithm (Algorithm~\ref{Alg:PairwiseAlg}) that finds these node neighborhoods $\hat{\mathcal{N}}_n(\svert) = \text{Supp}(\widehat{\Theta}_\svert)$ of each random variable $X_\svert$ separately by a greedy stagewise optimization of the conditional log-likelihood of $X_\svert$ conditioned on $X_{\vertex\backslash\svert}$. The algorithm then combines these neighborhoods to obtain a graph estimate $\widehat{E}$ using an ``OR'' rule: $\widehat{E}_n= \cup_\svert \{(\svert,t):\, t \in \hat{\mathcal{N}}_n(\svert) \}$. Other rules such as the ``AND'' rule, that add an edge only if it occurs in each of the respective node neighborhoods, could be used to combine the node-neighborhoods to a graph estimate. We show in Theorem~\ref{thr:pairwise} that the neighborhood selection by the greedy algorithm succeeds in recovering the exact node-neighborhoods with high probability, so that by a union bound, the graph estimates using either the AND or OR rules would be exact with high probability as well.

Before we describe this greedy algorithm and its analysis in Section~\ref{SecAlgo} however, we first consider the general statistical model case in the next section. We first describe the forward backward greedy algorithm of \citet{Zhang09} as applied to general statistical models, followed by a sparsistency analysis for this general case. We then specialize these general results in Section~\ref{SecAlgo} to the graphical
model case. The next section is thus of independent interest even outside the context of graphical models.

\section{Greedy Algorithm for General Losses}

Consider a random variable $Z$ with distribution $\mprob$, and let 
\mbox{$Z_{1}^{\numobs} \defn \{Z_1,\hdots,Z_\numobs\}$} denote 
$\numobs$ observations drawn i.i.d. according to $\mprob$. Suppose we are 
interested in estimating some parameter $\tparam \in \real^\pdim$ of the 
distribution $\mprob$ that is sparse; denote its number of non-zeroes by $s^* := \|\tparam\|_0$.
Let \mbox{$\Loss: \paramset \times \mathcal{Z}^\numobs \mapsto \real$} be some loss function that
assigns a cost to any parameter $\param \in \real^\pdim$, for a given set of observations $Z_{1}^{\numobs}$. 
For ease of notation, in the sequel, we adopt the shorthand $\Loss(\param)$ for $\Loss(\param; Z_{1}^{\numobs})$. We assume that $\tparam$ satisfies $\mathbb{E}_{Z}\left[\grad\Loss(\tparam)\right]=0$.

{\tiny
\begin{algorithm}[t]
\caption{\small Greedy forward-backward algorithm for finding a sparse optimizer of $\Loss(\cdot)$}
\label{Alg:General}
\begin{algorithmic}
\STATE {\small{\bf Input}: Data $\Data \defn \{x^{(1)}, \ldots, x^{(\numobs)}\}$, Stopping Threshold $\EpsilonStop$, Backward Step Factor $\nu \in (0,1)$}
\STATE {\small {\bf Output}: Sparse optimizer $\paramHat$}
\STATE
\STATE {\small $\paramHat^{(0)} \longleftarrow \mathbf{0}\,$ and $\suppHat^{(0)} \longleftarrow \phi\,$ and $\,k\longleftarrow 1$}
\STATE
\WHILE[\textit{Forward Step}]{true}
\STATE {\small $\displaystyle(j_*,\alpha_*)\longleftarrow\arg \min_{j\in\left(\suppHatkm\right)^c\,;\,\alpha}   
						\L(\paramHatkm\!+\!\alpha e_j;\Data)$}
\STATE {\small $\suppHatk \longleftarrow \suppHatkm \cup \{j_*\}$}\\
\STATE {\small $\delta_f^{(k)} \longleftarrow \Loss(\paramHatkm;\Data) - \Loss(\paramHatkm+\alpha_*e_{j_*};\Data)$}
\IF {\small $\delta_f^{(k)}\leq\EpsilonStop$}
\STATE {\small \bf break}
\ENDIF
\STATE
\STATE {\small $\displaystyle \paramHatk \longleftarrow 
		\arg\min_{\,\theta} \,\L\big(\theta_{\suppHatk};\Data\big)$}\\
\STATE {\small $k \longleftarrow k+1$}\\
\STATE
\WHILE[\textit{Backward Step}]{true}
\STATE {\small $\displaystyle j^* \longleftarrow \arg\min_{j \in \suppHatkm}\Loss(\paramHatkm-\paramHatkm_je_j;\Data)$}\\
\IF {\small $\Loss\big(\paramHatkm - \paramHatkm_{j^*}e_{j^*};\Data\big) - \Loss\big(\paramHatkm;\Data\big) > \nu\delta_f^{(k)}$}
\STATE {\small \bf break}
\ENDIF
\STATE
\STATE {\small $\suppHatkm\longleftarrow\suppHatk-\{j^*\}$}\\
\STATE {\small $\displaystyle \paramHatkm \longleftarrow 
		\arg\min_{\,\theta} \,\L\big(\theta_{\suppHatkm};\Data\big)$}\\
\STATE {\small $k\longleftarrow k-1$}\\
\ENDWHILE
\STATE
\ENDWHILE
\end{algorithmic}
\end{algorithm}}

We now consider the forward backward greedy algorithm in Algorithm \ref{Alg:General} that rewrites the algorithm in \cite{Zhang08} to allow for general loss functions. The algorithm starts with an empty set of active variables $\widehat{S}^{(0)}$ and gradually adds (and removes) vairables to the active set until it meets the stopping criterion. This algorithm has two major steps: the forward step and the backward step. In the forward step, the algorithm finds the \emph{best} next candidate and adds it to the active set as long as it improves the loss function at least by $\EpsilonStop$, otherwise the stopping criterion is met and the algorithm terminates. Then, in the backward step, the algorithm checks the \emph{influence} of all variables in the presence of the new added variable. If one or some of the previously added variables do not contribute at least $\nu\EpsilonStop$ to the loss function, then the algorithm removes them from the active set. This procedure ensures that at each round, the loss function is improved by at least $(1-\nu)\EpsilonStop$ and hence it terminates within a finite number of steps.

We state the assumptions on the loss function so that sparsistency could be guaranteed. Let us first recall the definition of 
restricted strong convexity from \citet{NRWY09}. Specifically, for a given set $\set$, the loss function is said to satisfy 
restricted strong convexity (RSC) with parameter $\LipshitzLower$ if
\begin{align}
\label{EqnDefnRSC}
\Loss(\param + \delpar; Z_{1}^{\numobs}) - \Loss(\param; Z_{1}^{\numobs})
    - \inprod{\nabla \Loss(\param; Z_{1}^{\numobs})}{\delpar} & \; \geq \; \frac{\LipshitzLower}{2} \,
    \|\delpar\|_2^2 \qquad \mbox{for all $\delpar \in \set$.}
\end{align}
We can now define sparsity restricted strong convexity as follows. Specifically, we say that the loss function $\Loss$ satisfies $RSC(k)$ with parameter $\LipshitzLower$
if it satisfies RSC with parameter $\LipshitzLower$ for all sets $S \subseteq \{1,\hdots,p\}$ such that $\|S\|_0 \le k$.

In contrast, we say the loss function satisfies restricted strong smoothness (RSS) with parameter $\LipshitzUpper$, for a given set $\set$ if 
\begin{align} \nonumber
\Loss(\param + \delpar; Z_{1}^{\numobs}) - \Loss(\param; Z_{1}^{\numobs})
    - \inprod{\nabla \Loss(\param; Z_{1}^{\numobs})}{\delpar} & \; \leq \; \frac{\LipshitzUpper}{2} \,
    \|\delpar\|_2^2 \qquad \mbox{for all $\delpar \in \set$.}
\end{align}
We can define $RSS(k)$ similarly: the loss function $\Loss$ satisfies $RSS(k)$ with parameter $\LipshitzUpper$ if it satisfies RSS with parameter $\LipshitzUpper$ for all sets 
$S \subseteq \{1,\hdots,p\}$ such that $\|S\|_0 \le k$ at all points $\param$ with $\|\param\|_0 \le k$. Given any constants $\LipshitzLower$ and $\LipshitzUpper$, and a sample based loss function $\Loss$, we can typically
use concentration based arguments to obtain bounds on the sample size required so that the $RSS$ and $RSC$ conditions hold with high probability.

Another property of the loss function that we require is an upper bound $\noiseLevel$ on the $\ell_\infty$ norm of the gradient of the loss at the true parameter $\tparam$, i.e., $\noiseLevel \ge \|\grad \Loss(\paramStar)\|_{\infty}$. This captures the ``noise level'' of the samples with respect to the loss. Here too, we can typically use concentration arguments to show for instance that $\noiseLevel \le \ConstantNoise (\log(p)/n)^{1/2}$, for some constant $\ConstantNoise > 0$ with high probability.
\vskip0.1in
\def\RSMult{\eta}
\begin{theorem}[Sparsistency] \label{thr:general}
Suppose the loss function $\Loss(\cdot)$ satisfies $RSC\left(\RSMult \, s^*\right)$ and $RSS\left(\RSMult \, s^*\right)$ with parameters $\LipshitzLower$ and $\LipshitzUpper$ for some 
{\small $\RSMult \ge 2 + 4 \rho^2 (\sqrt{(\rho^2 - \rho)/s^*} + \sqrt{2})^{2}$} with {\small $\rho=\LipshitzUpper/\LipshitzLower$}. Moreover, suppose that the true parameters $\paramStar$ satisfy 
$\min_{j \in \suppStar} |\paramStar_j| > \sqrt{32 \rho \EpsilonStop/\LipshitzLower}.$
Then if we run Algorithm~\ref{Alg:General} with stopping threshold $\EpsilonStop \ge (8 \rho \RSMult/\LipshitzLower) \; s^* \noiseLevel^2,$
the output $\paramHat$ with support $\suppHat$ satisfies: 
\begin{itemize}
	\item[(a)] {\bf Error Bound:} $\;\|\paramHat - \paramStar\|_2 \le \frac{2}{\LipshitzLower}\, \sqrt{s^*} \, (\noiseLevel \sqrt{\RSMult} + \sqrt{\EpsilonStop} \sqrt{2 \LipshitzUpper}).$
	\item[(b)] {\bf No False Exclusions:} $\suppStar - \suppHat = \emptyset.$
	\item[(c)] {\bf No False Inclusions:} $\suppHat - \suppStar = \emptyset.$
\end{itemize}	
\end{theorem}

\begin{proof}
The proof theorem hinges on three main lemmas: Lemmas~\ref{LemForwardStep} and \ref{LemBackwardStep} are simple consequences of the forward and backward steps failing when 
the greedy algorithm stops, and Lemma~\ref{LemErrorBound} which uses these two lemmas and extends techniques from \cite{Rothman2007} and \cite{NRWY10} to obtain an $\ell_2$ error bound on the error. Provided these lemmas hold, we then show below that the greedy algorithm is sparsistent. However, these lemmas require \emph{apriori} that the RSC and RSS conditions hold for sparsity size $|\suppStar \union \suppHat|$.
Thus, we use the result in Lemma~\ref{LemStoppingSize} that if $RSC(\RSMult s^*)$ holds, then the solution when the algorithm terminates satisfies $|\suppHat| \le (\RSMult - 1)s^*$, and hence $|\suppHat \union \suppStar|\le \RSMult s^*$. Thus, we can then apply Lemmas~\ref{LemForwardStep}, \ref{LemBackwardStep} and Lemma~\ref{LemErrorBound} to complete the proof as detailed below.
\begin{list}{\labelitemi}{\leftmargin=1em}
\item[(a)] The result follows directly from Lemma~\ref{LemErrorBound}, and noting that $|\suppHat \union \suppStar|\le \RSMult s^*$. In that Lemma, we show that the upper bound
holds by drawing from fixed point techniques in \cite{Rothman2007} and \cite{NRWY10}, and by using a simple consequence of the forward step failing when the greedy algorithm stops.

\item [(b)] Following the argument in \cite{Zhang08}, we use the chaining argument. For any $\tau\in\real$, we have
\small\begin{align*}
	\tau |\{j \in \suppStar - \suppHat : |\paramStar_j|^2  > \tau\}| &\le \|\paramStar_{\suppStar - \suppHat}\|_2^2 \,\le \|\paramStar - \paramHat\|_2^2\\
			&\le \frac{8 \RSMult s^*\noiseLevel^2}{\LipshitzLower^2} + \frac{16 \LipshitzUpper \EpsilonStop}{\LipshitzLower^2}\, |\suppStar - \suppHat|,
\end{align*}\normalsize
where the last inequality follows from part (a) and the inequality $(a+b)^2 \leq 2a^2+2b^2$. Now, setting $\tau = \frac{32 \LipshitzUpper \EpsilonStop}{\LipshitzLower^2}$, and dividing both sides by $\tau/2$ we get
\small\begin{align*}
	2 |\{j \in \suppStar - \suppHat : |\paramStar_j|^2  > \tau\}| &\le \frac{\RSMult s^* \noiseLevel^2}{2 \LipshitzUpper \EpsilonStop} + |\suppStar - \suppHat|.
\end{align*}\normalsize
Substituting \small$|\{j \in \suppStar - \suppHat : |\paramStar_j|^2  > \tau\}| = |\suppStar - \suppHat| - |\{j \in \suppStar - \suppHat : |\paramStar_j|^2  \le \tau\}| $\normalsize, we get
\small\begin{align*}
	|\suppStar - \suppHat| &\le |\{j \in \suppStar - \suppHat : |\paramStar_j|^2  \le \tau\}| + \frac{\RSMult s^* \noiseLevel^2}{2 \LipshitzUpper \EpsilonStop} \;\le |\{j \in \suppStar - \suppHat : |\paramStar_j|^2  \le \tau\}| + 1/2,
\end{align*}\normalsize
due to the setting of the stopping threshold $\EpsilonStop$. This in turn entails that 
\begin{align*}
	|\suppStar - \suppHat| \le |\{j \in \suppStar - \suppHat : |\paramStar_j|^2  \le \tau\}| = 0,
\end{align*}
by our assumption on the size of the minimum entry of $\paramStar$.

\item [(c)] From Lemma~\ref{LemBackwardStep}, which provides a simple consequence of the backward step failing when the greedy algorithm stops, for $\delparHat=\paramHat - \paramStar$, we have
$\EpsilonStop/\LipshitzUpper |\suppHat - \suppStar| \le \|\delparHat_{\suppHat - \suppStar}\|_2^2 \le \|\delparHat\|_2^2,$
so that using Lemma~\ref{LemErrorBound} and that $|\suppStar - \suppHat| = 0$, we obtain that 
\small$|\suppHat - \suppStar| \le \frac{4 \RSMult s^* \noiseLevel^2\LipshitzUpper}{\EpsilonStop \LipshitzLower^2} \le 1/2$\normalsize,
due to the setting of  the stopping threshold $\EpsilonStop$.
\end{list}

\vspace{-0.5cm}
\end{proof}


\subsection{Lemmas for Theorem~\ref{thr:general}}
We list the simple lemmas that characterize the solution obtained when the algorithm terminates, and on which the proof of Theorem~\ref{thr:general} hinges.
\begin{lemma}[Stopping Forward Step] When the algorithm \ref{Alg:General} stops with parameter $\paramHat$ supported on $\suppHat$, we have
\label{LemForwardStep}
\small\begin{align} \nonumber
	\left|\Loss\left(\paramHat\right) - \Loss\left(\paramStar\right)\right| < \sqrt{2\,|\suppStar - \suppHat| \, \LipshitzUpper\, \EpsilonStop} \; \left\|\paramHat - \paramStar\right\|_2. 
\end{align}\normalsize
\end{lemma}

\begin{lemma}[Stopping Backward Step]
\label{LemBackwardStep}
When the algorithm \ref{Alg:General} stops with parameter $\paramHat$ supported on $\suppHat$, we have
\small\begin{align}\nonumber
	\left\|\delparHat_{\suppHat - \suppStar}\right\|_2^2 \ge \frac{\EpsilonStop}{\LipshitzUpper} \left|\suppHat - \suppStar\right|.
\end{align}\normalsize
\end{lemma}

\begin{lemma}[Stopping Error Bound] When the algorithm  \ref{Alg:General} stops with parameter $\paramHat$ supported on $\suppHat$, we have
\label{LemErrorBound}
\small\begin{align}\nonumber
	\left\|\paramHat - \paramStar\right\|_2 \le \frac{2}{\LipshitzLower} \left(\noiseLevel \sqrt{\left|\suppStar \union \suppHat\right|} + \sqrt{2\left|\suppStar - \suppHat\right| \LipshitzUpper \EpsilonStop}\right).
\end{align}	\normalsize
\end{lemma}

\begin{lemma}[Stopping Size]
\label{LemStoppingSize}
If \small$\EpsilonStop>\frac{\noiseLevel^2}{\LipshitzUpper} \left(\sqrt{\frac{2}{\RSMult-1}} - \sqrt{\frac{2}{\RSMult}}\right)^{\!\!-2}\,$\normalsize and $RSC\left(\RSMult s^*\right)$ holds for some \small$\RSMult\geq 2 + 4\rho^2\left(\sqrt{\frac{\rho^2-\rho}{s^*}}+\sqrt{2}\right)^{\!\!2}\,$\normalsize, then the algorithm \ref{thr:general} stops with $k\leq (\RSMult-1) s^*$.
\end{lemma}

Notice that if $\EpsilonStop \ge (8 \rho \RSMult/\LipshitzLower) \; (\RSMult^2/(4\rho^2)) \; \noiseLevel^2,$ then, the assumption of this lemma is satisfied. Hence for large value of $s^*\geq 8 \rho^2 > \RSMult^2/(4\rho^2)$, it suffices to have $\EpsilonStop \ge (8 \rho \RSMult/\LipshitzLower) \; s^* \noiseLevel^2$.

\section{Greedy Algorithm for Pairwise Graphical Models}\label{SecAlgo}

Suppose we are given set of $n$ i.i.d. samples $\Data \defn \{x^{(1)}, \ldots, x^{(\numobs)}\}$, drawn from a pairwise Ising model as in \eqref{EqnIsing}, with parameters $\tparam$,
and graph $G = (V,E^*)$. It will be useful to denote the maximum node-degree in the graph $E^*$ by $d$. As we will show, our model selection performance depends critically on this parameter $d$.
We then propose the Algorithm~\ref{Alg:PairwiseAlg} for estimating the underlying graphical model from the $n$ samples $\Data$. 

\begin{algorithm}[t]
\caption{\small Greedy forward-backward algorithm for pairwise discrete graphical model learning}
\label{Alg:PairwiseAlg}
\begin{algorithmic}
\STATE {\small {\bf Input}: Data $\Data \defn \{x^{(1)}, \ldots, x^{(\numobs)}\}$, Stopping Threshold $\EpsilonStop$, Backward Step Factor $\nu \in (0,1)$}
\STATE {\small {\bf Output}: Estimated Edges $\widehat{E}$}
\STATE
\FOR{$\svert\in\vertex$}
\STATE {\small Run Algorithm~\ref{Alg:General} with $\Loss(\cdot)$ described by \eqref{eq:Loss-Fn} to get $\Theta_\svert$ and its support $\widehat{\Nr}$}
\ENDFOR
\STATE
\STATE {\small Output $\widehat{E}=\bigcup_\svert\left\{(\svert,t):t\in\widehat{\Nr}\right\}$}
\end{algorithmic}
\end{algorithm}

\begin{theorem}[Pairwise Sparsistency] \label{thr:pairwise}

Suppose we run Algorithm~\ref{Alg:PairwiseAlg} with stopping threshold $\EpsilonStop \ge c_1 \frac{d\, \log p}{n}$, where, $d$ is the maximum node degree in the graphical model, and the true parameters $\paramStar$ satisfy 
$\frac{c_3}{\sqrt{d}} > \min_{j \in \suppStar} |\paramStar_j| > c_2 \sqrt{\EpsilonStop}$,
and further that number of samples scales as $$n > c_4\, d^2\, \log p,$$ for some constants $c_1,c_2,c_3,c_4$. Then, with probability at least $1 - c' \exp(- c'' n)$, the output $\paramHat$ supported on $\suppHat$ satisfies: 
\begin{itemize}
	\item[(a)] {\bf No False Exclusions:} $E^* - \widehat{E} = \emptyset.$
	\item[(b)] {\bf No False Inclusions:} $\widehat{E} - E^* = \emptyset.$
\end{itemize}	
\end{theorem}

\def\KappaOneUpper{\kappa_{1}^{u}}
\def\KappaOneLower{\kappa_{1}^{l}}
\def\KappaTwoUpper{\kappa_{2}^{u}}
\def\KappaTwoLower{\kappa_{2}^{l}}

\begin{proof}
This theorem is a corollary to our general Theorem~\ref{thr:general}. We first show that the conditions of Theorem~\ref{thr:general} hold under the assumptions in this corollary. 

\myparagraph{RSC, RSS}
We first note that the conditional log-likelihood loss function in \eqref{eq:Loss-Fn} corresponds to a logistic likelihood. Moreover, the covariates are all binary, and bounded, and hence also sub-Gaussian. \cite{NRWY10,ANW10} analyze the RSC and RSS properties of generalized linear models, of which logistic models are an instance, and show that the following result holds if the covariates are sub-Gaussian. Let $\partial \Loss(\delpar; \tparam) = \Loss(\tparam + \delpar) - \Loss(\tparam)  - \inprod{\nabla \Loss(\tparam)}{\delpar}$ be the second order Taylor series remainder. Then, Proposition~2 in \cite{NRWY10} states that that there exist constants $\KappaOneLower$ and $\KappaTwoLower$, independent of $n,p$ such that with probability at least $1 - c_1 \exp( - c_2 n)$, for some constants $c_1,c_2 > 0$,
\begin{align*}
\partial \Loss(\delpar; \tparam) & \; \geq \; \KappaOneLower \|\delpar\|_2 \left\{ \|\delpar\|_2 - \KappaTwoLower \sqrt{\frac{\log (p)}{n}} \|\delpar\|_1\right\} \qquad \mbox{for all $\delpar: \|\delpar\|_2 \le 1$}.
\end{align*}
Thus, if $\|\delpar\|_0 \le k\defn \eta d$, then $\|\delpar\|_1 \le \sqrt{k} \|\delpar\|_2$, so that 
\begin{align*}
	\partial \Loss(\delpar; \tparam) & \; \geq \; \|\delpar\|^2_2 \left(\KappaOneLower  - \KappaTwoLower \sqrt{\frac{k \log p}{n}} \right)\; \geq \; \frac{\KappaOneLower}{2} \|\delpar\|^2_2,
\end{align*}
if $n > 4 (\KappaTwoLower/\KappaOneLower)^2 \, \eta d\, \log (p)$. In other words, with probability at least $1 - c_1 \exp( - c_2 n)$, the loss function $\Loss$ satisfies $RSC(k)$ with parameter $\KappaOneLower$ provided $n > 4 (\KappaTwoLower/\KappaOneLower)^2 \, \eta d\, \log (p)$.
Similarly, it follows from \cite{NRWY10,ANW10} that there exist constants $\KappaOneUpper$ and $\KappaTwoUpper$ such that with probability at least $1 - c'_1 \exp( - c' _2 n)$,
\begin{align*}
\partial \Loss(\delpar; \tparam) & \; \leq \; \KappaOneUpper \|\delpar\|_2 \{ \|\delpar\|_2 - \KappaTwoUpper \|\delpar\|_1\} \qquad \mbox{for all $\delpar: \|\delpar\|_2 \le 1$},
\end{align*}
so that by a similar argument, with probability at least $1 - c'_1 \exp( - c'_2 n)$, the loss function $\Loss$ satisfies $RSS(k)$ with parameter $\KappaOneUpper$ provided $n > 4 (\KappaTwoUpper/\KappaOneUpper)^2 \, \eta d\, \log (p)$.

\myparagraph{Noise Level}
Next, we obtain a bound on the noiselevel $\noiseLevel \ge \|\grad \Loss(\tparam)\|_{\infty}$ following similar arguments to \cite{RWLIsing}. Let $W$ denote the gradient $\grad \Loss(\tparam)$ of the loss function~\eqref{eq:Loss-Fn}. Any entry of $W$ has the form $W_{t} = \frac{1}{n}\sum_{i=1}^{n} Z^{(i)}_{rt}$, where $Z^{(i)}_{rt} = x^{(i)}_{t} ( x^{(i)}_r - \mathbb{P}(x_r = 1|x_{\backslash s}^{(i)}))$
are zero-mean, i.i.d. and bounded $|Z^{(i)}_{rt}| \le 1$. Thus, an application of Hoeffding's inequality yields that $\mathbb{P}[|W_{t}| > \delta] \le 2 \exp(- 2n \delta^2).$ Applying a union bound over indices in $W$, we get $\mathbb{P}[\|W\|_\infty > \delta] \le 2 \exp(- 2n \delta^2 + \log (p))$. Thus, if $\lambda_n = (\log (p) /n)^{1/2}$, then $\|W\|_{\infty} \le \lambda_n$
with probability at least $1 - \exp(-n \lambda_n^2+\log(p))$.

We can now verify that under the assumptions in the corollary, the conditions on the stopping size $\EpsilonStop$ and the minimum absolute value of the non-zero parameters $\min_{j \in \suppStar} |\paramStar_j|$ are satisfied. Moreover, from the discussion above, under the sample size scaling in the corollary, the required $RSC$ and $RSS$ conditions hold as well. Thus, Theorem~\ref{thr:general} yields
that each node neighborhood is recovered with no false exclusions or inclusions with probability at least $1 - c' \exp(- c'' n)$. An application of a union bound over all nodes completes the proof.

\end{proof}

{\bf \it Remarks.} The sufficient condition on the parameters imposed by the greedy algorithm is a restricted strong convexity condition~\cite{NRWY10}, which is weaker than the irrepresentable condition required by \cite{RWLIsing}. Further, the number of samples required for sparsistent graph recovery scales as $O(d^2 \log p)$, where $d$ is the maximum node degree, in contrast to $O(d^3 \log p)$ for the $\ell_1$ regularized counterpart. We corroborate this in our simulations, where we find that the greedy algorithm requires fewer observations than \cite{RWLIsing} for sparsistent graph recovery.

We also note that the result can also be extended to the general pairwise graphical model case, where each random variable takes values in the range $\{1,\ldots,m\}$. In that case, 
the conditional likelihood of each node conditioned on the rest of the nodes takes the form of a multiclass logistic model, and the greedy algorithm would take the form of 
a ``group'' forward-backward greedy algorithm, which would add or remove all the parameters corresponding to an edge as a group. Our analysis however naturally extends to such a 
group greedy setting as well. The analysis for RSC and RSS remains the same and for bounds on $\noiseLevel$, see equation (12) in \cite{JRVS11}. We defer further discussion on this due to the lack of space.

\section{Experimental Results}\label{SecExper}
We now present experimental results that illustrate the power of Algorithm~\ref{Alg:PairwiseAlg} and support our theoretical guarantees.  We simulated structure learning of several different graph structures and compared the learning rates of our method against that of a standard $\ell_1$-logistic regression method as outlined in \cite{RWLIsing}.    

We performed experiments using 3 different graph structures: (a) chain (line graph), (b) 4-nearest neighbor (grid graph) and (c) star graph. For each experiment, we assumed a pairwise binary Ising model in which each $\theta_{\svert t}^{*}=\pm 1$ randomly.  For each graph type, we generated a set of $n$ samples ${x^{(1)},...,x^{(n)}}$ using Gibbs sampling.  We then attempted to learn the structure of the model using both Algorithm \ref{Alg:PairwiseAlg} as well as $\ell_1$-logistic regression.  We then compared the actual graph structure with the empirically learned graph structures.  If the graph structures matched completely then we declared the result a \emph{success} otherwise we declared the result a \emph{failure}.  We compared these results over a range of sample sizes ($n$) and averaged the results for each sample size over a batch of size $10$.  For all greedy experiments we set the stopping threshold $\EpsilonStop=\frac{c\log(np)}{n} $, where $c$ is a tuning constant, as suggested by Theorem~\ref{thr:pairwise}, and set the backwards step threshold $\nu=0.5$.  For all $\ell_1$-logistic regression experiments we set the regularization parameter $\lambda_n=c'\sqrt{\log(p)/n}$, where $c'$ is set via cross-validation.

\begin{figure}[t]
	\renewcommand{\figurename}{Fig}
	\centering
	\subfigure[Chain (Line Graph)]{\includegraphics[width=0.48\textwidth]{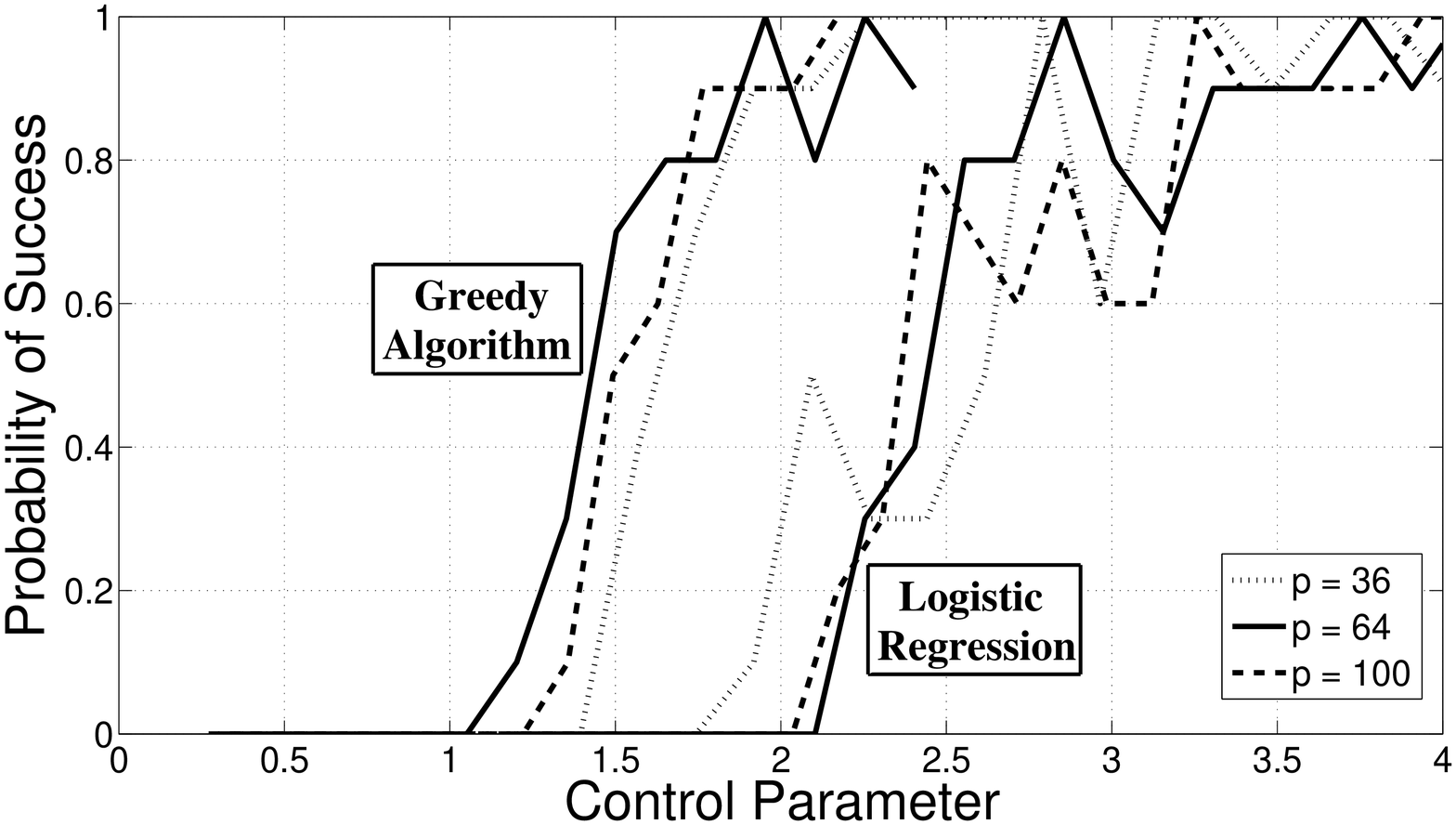}}
	\subfigure[4-Nearest Neighbor (Grid Graph)]{\includegraphics[width=0.48\textwidth]{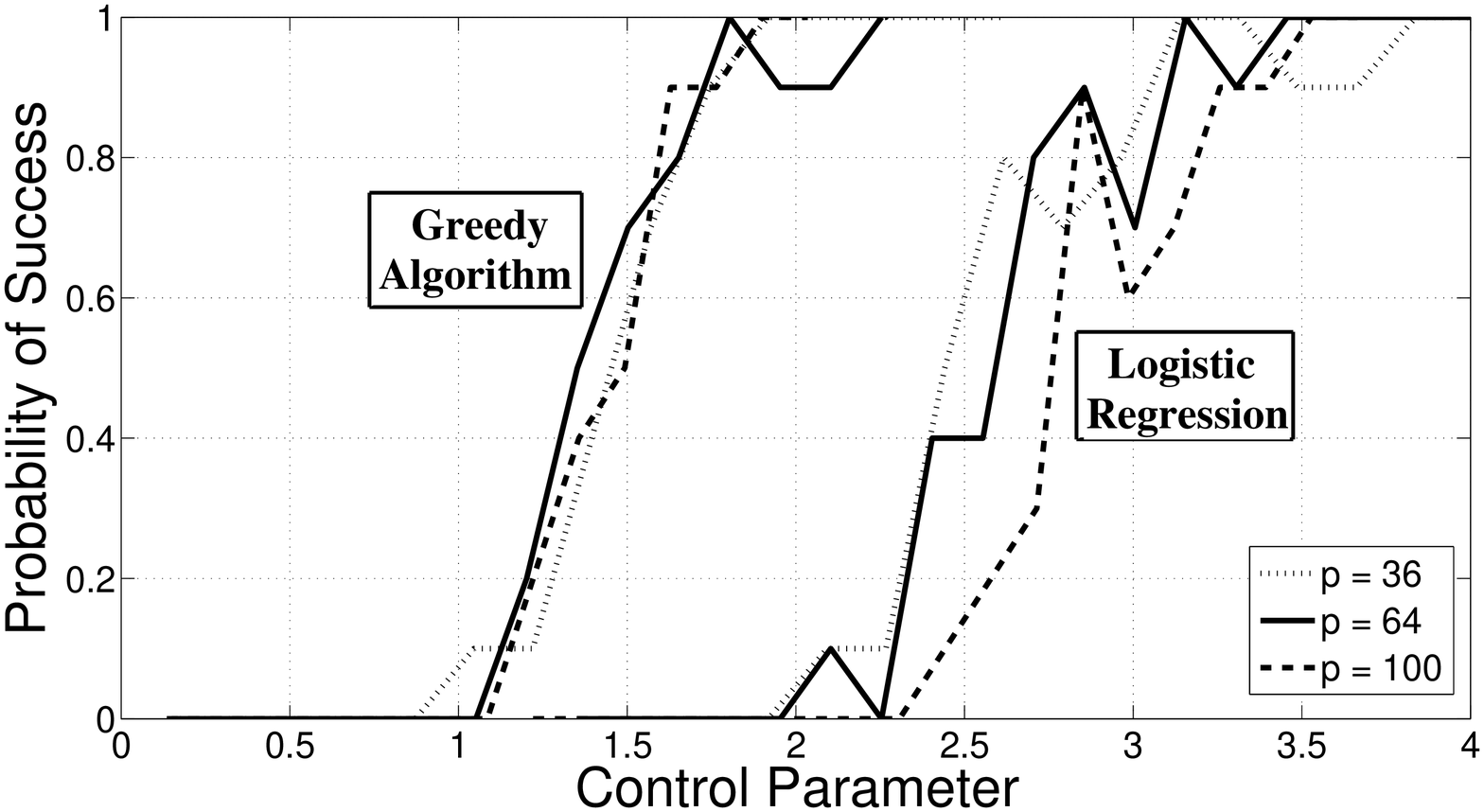}}
	\subfigure[Star]{\includegraphics[width=0.48\textwidth]{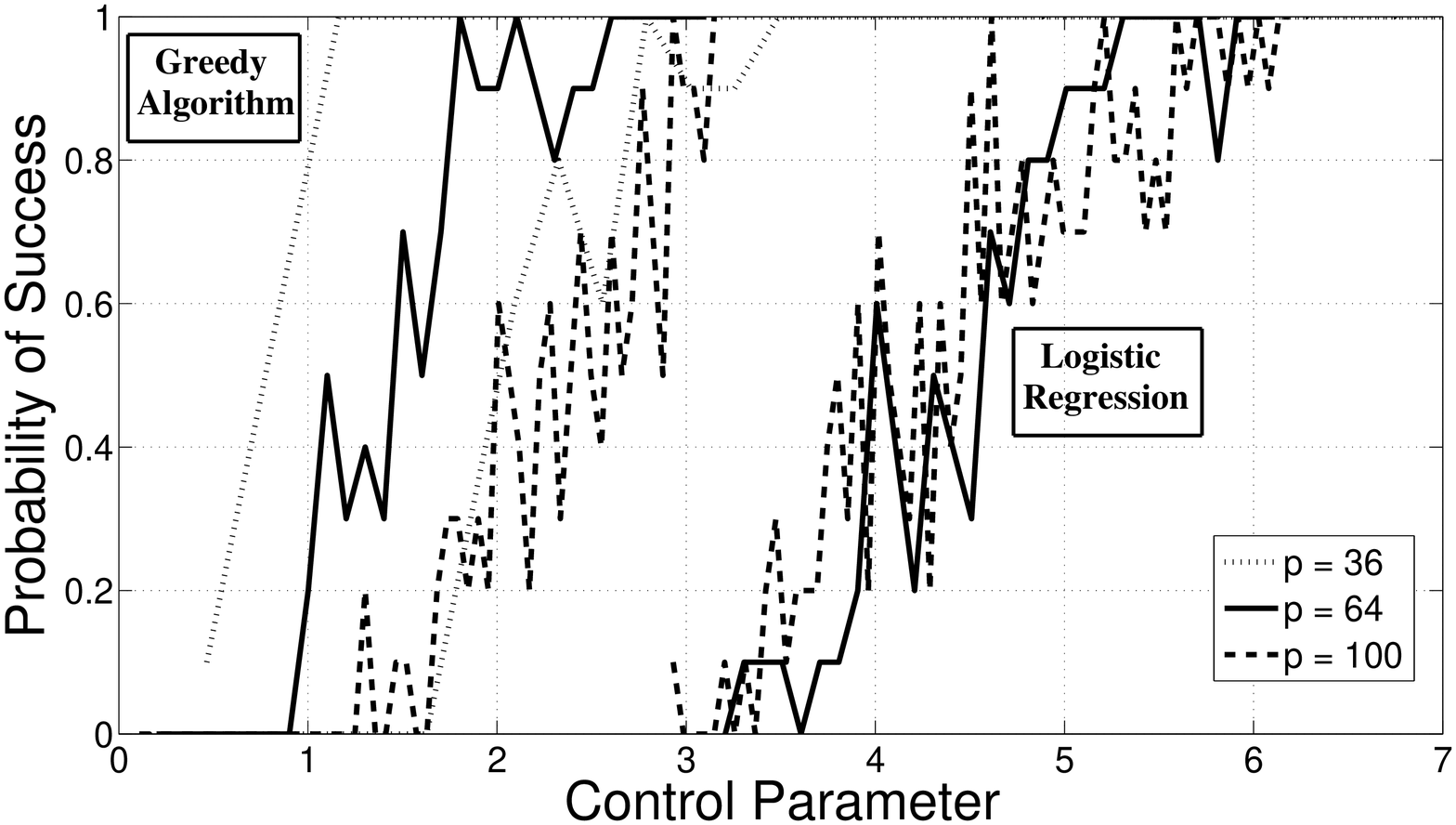}}
	\subfigure[Chain, 4-Nearest Neighbor and Star Graphs]{\includegraphics[width=0.48\textwidth]{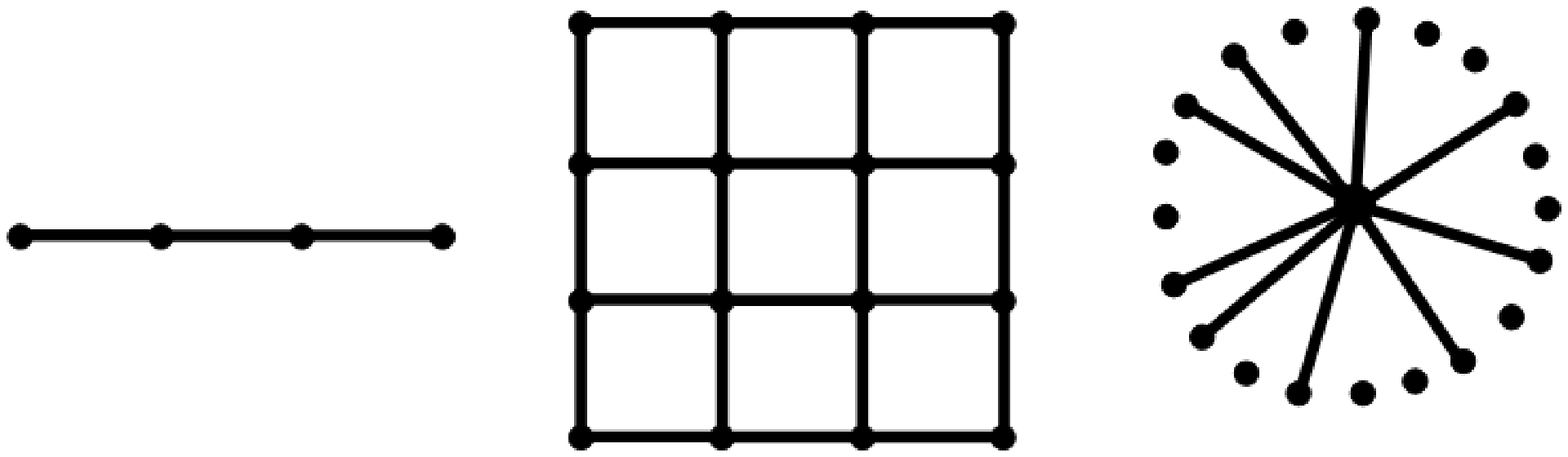}}
	\caption{Plots of success probability $\mathbb{P}[\widehat{\mathcal{N}}_{\pm}(\svert)=\mathcal{N}^*(\svert),\forall \svert\in\vertex]$ versus the control parameter $\beta(n,p,d)=n/[20d\log(p)]$ for Ising model on (a) chain $(d=2)$, (b) 4-nearest neighbor $(d=4)$ and (c) Star graph $(d=0.1p)$. The coupling parameters are chosen randomly from $\theta_{st}^{*}=\pm0.50$ for both greedy and $\ell_1$-logistic regression methods. As our theorem suggests and these figures show, the greedy algorithm requires less samples to recover the exact structure of the graphical model.}
	\label{fig:plots}
\end{figure}

Figure~\ref{fig:plots} shows the results for the chain $(d=2)$, grid $(d=4)$ and star $(d=0.1p)$ graphs using both Algorithm~\ref{Alg:PairwiseAlg} and $\ell_1$-logistic regression for three different graph sizes $p\in\{36,64,100\}$ with mixed (random sign) couplings. For each sample size, we generated a batch of $10$ different graphical models and averaged the probability of success (complete structure learned) over the batch. Each curve then represents the probability of success versus the control parameter $\beta(n,p,d)=n/[20d\log(p)]$ which increases with the sample size $n$. These results support our theoretical claims and demonstrate the efficiency of the greedy method in comparison to node-wise logistic regression \cite{RWLIsing}.

\newpage
\small
\bibliographystyle{plainnat}
\bibliography{GreedyGM}
\normalsize

\newpage
\appendix

\section{Auxiliary Lemmas for Theorem~1}
In this section, we prove the Lemmas used in the proof of Theorem 1. Note that when the algorithm terminates, the forward step fails to go through. This entails that 
\begin{align}
\Loss(\paramHat) - \inf_{j \in \suppHat^c,\alpha \in \real} \Loss(\paramHat + \alpha e_j) < \EpsilonStop.
\end{align}

The next lemma shows that this has the consequence of upper bounding the deviation in loss between the estimated parameters $\paramHat$ and the true parameters $\paramStar$.

\begin{lemma}[Stopping Forward Step] When the algorithm stops with parameter $\paramHat$ supported on $\suppHat$, we have
\label{LemForwardStep}
\begin{align}
	\left|\Loss\left(\paramHat\right) - \Loss\left(\paramStar\right)\right| < \sqrt{2\,|\suppStar - \suppHat| \, \LipshitzUpper\, \EpsilonStop} \; \left\|\paramHat - \paramStar\right\|_2. 
\end{align}
\end{lemma}
\begin{proof}
Let $\widehat{\delpar} = \paramStar - \paramHat$. For any $\eta\in\real$, we have
\begin{align*}
	\Loss\left(\paramHat + \eta \widehat{\delpar}_j e_j\right) \le \Loss\left(\paramHat\right) + \eta \grad_j \Loss\left(\paramHat\right) \widehat{\delpar}_j + \eta^2 \frac{\LipshitzUpper}{2} \widehat{\delpar}_j^2.
\end{align*}	
Thus, we can establish
\begin{align*}	
	- |\suppStar - \suppHat| \EpsilonStop &< \sum_{j \in \suppStar - \suppHat} \Big(\Loss\left(\paramHat + \eta \widehat{\delpar}_j e_j\right) - \Loss\left(\paramHat\right)\Big)\\
									&\le \eta \Big(\Loss\left(\paramStar\right) - \Loss\left(\paramHat\right)\Big) + \eta^2 \frac{\LipshitzUpper}{2} \left\|\widehat{\delpar}\right\|_2^2.
\end{align*}
Optimizing the RHS over $\eta$, we obtain
\begin{align*}
	-|\suppStar - \suppHat|\, \EpsilonStop &< -\, \frac{\Big(\Loss(\paramStar) - \Loss\left(\paramHat\right)\Big)^2}{2\, \LipshitzUpper\, \|\widehat{\delpar}\|_2^2},
\end{align*}
whence the lemma follows.\\
\end{proof}

\begin{lemma}[Stopping Error Bound] When the algorithm stops with parameter $\paramHat$ supported on $\suppHat$, we have
\label{LemErrorBound}
\begin{align}
	\|\paramHat - \paramStar\|_2 \le \frac{2}{\LipshitzLower} \left(\noiseLevel \sqrt{\left|\suppStar \union \suppHat\right|} + \sqrt{2\left|\suppStar - \suppHat\right| \LipshitzUpper \EpsilonStop}\right).
\end{align}	
\end{lemma}
\begin{proof}
For $\delpar\in\real$, let 
\begin{align*}
	G(\delpar) &= \Loss\left(\paramStar + \delpar\right) - \Loss\left(\paramStar\right) - \sqrt{2\left|\suppStar - \suppHat\right| \, \LipshitzUpper \, \EpsilonStop} \left\|\delpar\right\|_2.
\end{align*}
It can be seen that $G(0) = 0$, and from the previous lemma, $G(\delparHat) \le 0$. Further, $G(\delpar)$ is sub-homogeneous (over a limited range): $G(t \delpar) \le t G(\delpar)$ for $t \in [0,1]$. Thus, for a carefully chosen $r > 0$, if we show that $G(\delpar) > 0$ for all $\delpar \in \{\delpar : \|\delpar\|_{2} \le r,\;\|\delpar\|_0 \le |\suppUnion|\}$, where $\suppUnion = |\suppHat \union \suppStar|$, then it follows that $\|\delparHat\|_2 \le r$.
If not, then there would exist some $t \in [0,1)$ such that $\|t \delparHat\| = r$, whence we would arrive at the contradiction
\begin{align*}
	0 < G(t \delparHat) \le t G(\delparHat) \le 0.
\end{align*}

Thus, it remains to show that $G(\delpar) > 0$ for all $\delpar \in \{\delpar : \|\delpar\|_{2} \le r,\;\|\delpar\|_0 \le |\suppUnion|\}$. By restricted strong convexity property of $\Loss$, we have
\begin{align*}
	\Loss(\paramStar + \delpar) - \Loss(\paramStar) \ge \tr{\grad \Loss(\paramStar)}{\delpar} + \frac{\LipshitzLower}{2} \left\|\delpar\right\|_2^2.
\end{align*}
We can establish
\begin{align*}
	\tr{\grad \Loss(\paramStar)}{\delpar} &\ge  - \left|\tr{\grad \Loss(\paramStar)}{\delpar}\right|\\
											&\ge - \left\|\grad \Loss(\paramStar)\right\|_\infty \left\|\delpar\right\|_1 = \noiseLevel \left\|\delpar\right\|_1,
\end{align*}
and hence,
\begin{align*}
G(\paramStar + \delpar) &\ge - \noiseLevel \|\delpar\|_1 + \frac{\LipshitzLower}{2} \|\delpar\|_2^2 - \sqrt{2 \left|\suppStar - \suppHat\right| \LipshitzUpper \EpsilonStop } \|\delpar\|_2\\
						&> \|\delpar\|_2 \left(  \frac{\LipshitzLower}{2} \|\delpar\|_2 - \noiseLevel \sqrt{\left|\suppStar \union \suppHat\right|} - \sqrt{2\left|\suppStar - \suppHat\right| \LipshitzUpper \EpsilonStop}  \right)\\
						&> 0,
\end{align*}
if $\|\delpar\|_2 = r$ for $$r = \frac{2}{\LipshitzLower} \left(\noiseLevel \sqrt{\left|\suppStar \union \suppHat\right|} + \sqrt{2 \left|\suppStar - \suppHat\right| \LipshitzUpper \EpsilonStop}\right).$$

This concludes the proof of the lemma.\\
\end{proof}

Next, we note that when the algorithm terminates, the backward step with the current parameters has failed to go through. This entails that 
\begin{align}\label{EqnBackwardStop}
\inf_{j \in \suppHat} \Loss(\paramHat - \paramHat_j e_j) - \Loss(\paramHat) > \EpsilonStop/2.
\end{align}
The next lemma shows the consequence of this bound.
\begin{lemma}[Stopping Backward Step]
\label{LemBackwardStep}
When the algorithm stops with parameter $\paramHat$ supported on $\suppHat$, we have
\begin{align}
	\left\|\delparHat_{\suppHat - \suppStar}\right\|_2^2 \ge \frac{\EpsilonStop}{\LipshitzUpper} \left|\suppHat - \suppStar\right|.
\end{align}
\end{lemma}
\begin{proof}
We have
\begin{align*}
	 |\suppHat - \suppStar|  \inf_{j \in \suppHat} \Loss(\paramHat - \paramHat_j e_j) &\le \sum_{j \in \suppHat - \suppStar} \Loss(\paramHat - \paramHat_j e_j)\\
						&\le |\suppHat - \suppStar| \Loss(\paramHat) + 
					 \sum_{j \in \suppHat - \suppStar} \left( \grad_{j} \Loss(\paramHat) \; \paramHat_j + \frac{\LipshitzUpper}{2} \paramHat_j^2\right)\\
						&\le |\suppHat - \suppStar|  \Loss(\paramHat) +  \frac{\LipshitzUpper}{2} \left\|\delparHat_{\suppHat - \suppStar}\right\|_2^2,
\end{align*}
where the second inequality uses the fact that $[\grad \Loss(\paramHat)]_{\suppHat} = 0$. Substituting \eqref{EqnBackwardStop} above, the lemma follows.	
\end{proof}


\section{Lemmas on the Stopping Size}

\begin{lemma}
\label{LemStoppingSize}
If $\EpsilonStop>\frac{\noiseLevel^2}{\LipshitzUpper} \left(\frac{\frac{1}{2\rho}\sqrt{\gamma}-\sqrt{\frac{\rho^2-\rho}{k^*}}}{\sqrt{1+\gamma}} - \sqrt{\frac{2}{2+\gamma}}\right)^{\!\!-2}$ and $RSC\left((2+\gamma)k^*\right)$ holds for some $\gamma\geq 4\rho^2\left(\sqrt{\frac{\rho^2-\rho}{k^*}}+\sqrt{2}\right)^{\!\!2}$, then the algorithm stops with $k\leq(1+\gamma)k^*$.
\end{lemma}

\begin{proof}
Consider the first time the algorithm reaches $k=(1+\gamma)k^*+1$, then by Lemma~\ref{lem:backwardstep} and \ref{lem:errorbound}, we have
\begin{equation}
\begin{aligned}
\sqrt{\frac{k-1-k^*}{k-1}}\leq \sqrt{\frac{|\suppHatkm - \suppStar|}{|\suppHatkm \cup \suppStar|}}&\leq \frac{2\LipshitzUpper\sqrt{\LipshitzUpper(\LipshitzUpper-\LipshitzLower)}} {\LipshitzLower^2\sqrt{|\suppHatkm \cup \suppStar|}} + \frac{2\LipshitzUpper}{\LipshitzLower} \left(\frac{\noiseLevel}{\sqrt{\LipshitzUpper\EpsilonStop}}  + \sqrt{\frac{2|\suppStar - \suppHatkm|}{|\suppStar \cup \suppHatkm|}} \,\right)\\
&\leq \frac{2\frac{\LipshitzUpper}{\LipshitzLower} \sqrt{\left(\frac{\LipshitzUpper}{\LipshitzLower}\right)^2-\frac{\LipshitzUpper}{\LipshitzLower}}}{\sqrt{k-1}} + \frac{2\LipshitzUpper}{\LipshitzLower} \left(\frac{\noiseLevel}{\sqrt{\LipshitzUpper\EpsilonStop}}  + \sqrt{\frac{2k^*}{k+k^*-1}} \,\right).
\end{aligned}
\nonumber
\end{equation}
Hence, we get 
\begin{equation}
\frac{\frac{1}{2\rho}\sqrt{\gamma}-\sqrt{\frac{\rho^2-\rho}{k^*}}}{\sqrt{1+\gamma}} - \sqrt{\frac{2}{2+\gamma}} \leq \frac{\noiseLevel}{\sqrt{\LipshitzUpper\EpsilonStop}}.\\
\nonumber
\end{equation}
For $\gamma\,\geq\,4\rho^2\left(\sqrt{\frac{\rho^2-\rho}{k^*}}+\sqrt{2}\right)^{\!\!2}$, the LHS is positive and we arrive to a contradiction with the assumption on $\EpsilonStop$.\\
\end{proof}

When the algorithm reaches the support size of $k$ at the \emph{beginning} of the forward step, i.e., we added the $k^{th}$ variable to the support and the backward step did not remove any variable, let $\paramHatk$ denote the current parameter and $\suppHatk = \text{Supp}(\paramHatk)$ with $k=|\suppHatk|$. Let $\paramStar$ be the target parameter matrix (i.e., $\mathbb{E}\left[\grad\Loss(\paramStar)\right]=0$), with $\suppStar = \text{Supp}(\paramStar)$ and $\kStar = |\suppStar|$. Lemmas~\ref{lem:backwardstep}, \ref{lem:forwardstep} and \ref{lem:errorbound} follow along similar lines to their counterparts in 
Lemmas~\ref{LemBackwardStep}, \ref{LemForwardStep} and \ref{LemErrorBound} respectively: the latter held when the algorithm terminates, while the lemmas below hold at any iterate $\paramHatk$ where we have first added the $k^{th}$ variable to the support. We provide their detailed proofs for completeness.

\begin{lemma}[General Backward Step] 
The first time the algorithm reaches a support size of $k>k^*+4\left(\frac{\LipshitzUpper}{\LipshitzLower}\right)^{\!\!4}+1$ at the beginning of the forward step, assuming $RSC\left(|\suppHatk\cup\suppStar|\right)$ holds, we have
\begin{equation}
\left\|\paramHatkm_{\suppHatkm - \suppStar}\right\|_2^2\geq \left(\sqrt{\frac{|\suppHatkm - \suppStar|}{\LipshitzUpper}} -\frac{2\LipshitzUpper\sqrt{\LipshitzUpper-\LipshitzLower}}{\LipshitzLower^2}\right)^{\!\!\!2}\delta_f^{(k)}. 
\end{equation}
\label{lem:backwardstep}
\end{lemma}

\begin{proof}
Under the assumption of the lemma, the immediate previous backward step has not gone through and hence,
\begin{equation} \nonumber
\inf_{j\in\suppHatk - \suppStar}\Loss\left(\paramHatk - \paramHatk_j e_j\right) - \Loss\left(\paramHatk\right) \geq \frac{\delta_f^{(k)}}{2}.
\end{equation}
Consequently, we get
\begin{equation}
\begin{aligned}
|\suppHatkm - \suppStar|\frac{\delta_f^{(k)}}{2} &\leq \sum_{j\in\suppHatkm - \suppStar}\Loss(\paramHatk - \paramHatk_j e_j) - \Loss(\paramHatk)\\
&\leq \frac{\LipshitzUpper}{2}\left\|\paramHatk_{\suppHatkm - \suppStar}\right\|_2^2\\
&\leq \frac{\LipshitzUpper}{2}\left(\left\|\paramHatkm_{\suppHatkm - \suppStar}\right\|_2 + \left\|\Delta^{(k)}\right\|_2\right)^2,
\end{aligned}
\nonumber
\end{equation}
where, $\Delta^{(k)} = \paramHatk_{\suppHatkm}-\paramHatkm$. This entails that 
\begin{equation}
\begin{aligned}
\left(\sqrt{\frac{|\suppHatkm - \suppStar|}{\LipshitzUpper}\delta_f^{(k)}}-\left\|\Delta^{(k)}\right\|_2\right)^2 &\leq \left\|\paramHatkm_{\suppHatkm - \suppStar}\right\|_2^2.
\end{aligned}
\nonumber
\end{equation}
Thus, it suffices to show that $\left\|\Delta^{(k)}\right\|_2\leq \frac{2\LipshitzUpper}{\LipshitzLower^2}\sqrt{(\LipshitzUpper-\LipshitzLower)\delta_f^{(k)}}$.

\bigskip
From the forward step, we have
\begin{equation} \nonumber
\Loss\left(\paramHatkm\right) - \inf_{j\notin\suppHatkm,\alpha\in\real}\Loss\left(\paramHatkm + \alpha e_j\right) = \delta_f^{(k)}.
\end{equation}
Let $(j_*,\alpha_*\neq 0)$ be the optimizer of the equation above. Now, we have
\begin{equation}
\begin{aligned}
\frac{\LipshitzLower}{2}\left\|\Delta^{(k)}\right\|_2^2 &\leq \Loss\left(\paramHatk_{\suppHatkm}\right)-\Loss\left(\paramHatkm\right)\\
&\leq \Loss\left(\paramHatk_{\suppHatkm}\right)-\Loss\left(\paramHatk\right) +\Loss\left(\paramHatk\right)-\Loss\left(\paramHatkm\right)\\
&\leq \frac{\LipshitzUpper}{2}\left|\paramHatk_{j_*}\right|^2 -\frac{\LipshitzLower}{2}\left\|\Delta^{(k)}\right\|_2^2 -\frac{\LipshitzLower}{2}\left|\paramHatk_{j_*}\right|^2.
\end{aligned}
\nonumber
\end{equation}
Hence, $\left\|\Delta^{(k)}\right\|_2^2\leq \frac{\LipshitzUpper-\LipshitzLower}{2\LipshitzLower}\left|\paramHatk_{j_*}\right|^2$ and we only need to show that $\left|\paramHatk_{j_*}\right|\leq\frac{2\LipshitzUpper}{\LipshitzLower}\sqrt{\frac{2}{\LipshitzLower}\delta_f^{(k)}}$. Since $\left|\paramHatk_{j_*}\right|\leq \left|\paramHatk_{j_*}-\alpha_*\right|+\left|\alpha_*\right|$, we can equivalently control the latter two terms. First, by forward step construction, $\frac{\LipshitzLower}{2}\left|\alpha^*\right|^2\leq \Loss\left(\paramHatkm\right)-\Loss\left(\paramHatkm+\alpha_*e_{j_*}\right)=\delta_f^{(k)}$ and hence $\left|\alpha^*\right|\leq \sqrt{\frac{2}{\LipshitzLower}\delta_f^{(k)}}$. Second, we claim that $\left|\paramHatk_{j_*}-\alpha_*\right|\leq \frac{2\LipshitzUpper-\LipshitzLower}{\LipshitzLower}\left|\alpha_*\right|$ and we are done. 

\bigskip
In contrary, suppose $\left|\paramHatk_{j_*}-\alpha_*\right|^2> \left(\frac{2\LipshitzUpper-\LipshitzLower}{\LipshitzLower}\right)^2\left|\alpha_*\right|^2\geq \frac{\LipshitzUpper}{\LipshitzLower}\left|\alpha_*\right|^2$. We have
\begin{equation}
\begin{aligned}
\frac{\LipshitzLower}{2}\left|\paramHatk_{j_*}-\alpha_*\right|^2 &> \frac{\LipshitzUpper}{2}\left|\alpha_*\right|^2\\ &\geq \Loss\left(\paramHatk-\alpha_*e_{j_*}\right) - \Loss\left(\paramHatk\right)\\ &\geq \Loss\left(\paramHatk-\alpha_*e_{j_*}\right) - \Loss\left(\paramHatkm\right) +\Loss\left(\paramHatkm\right) -\Loss\left(\paramHatk\right)\\ &\geq \frac{\LipshitzLower}{2}\left\|\Delta^{(k)}\right\|_2^2 + \frac{\LipshitzLower}{2}\left|\paramHatk_{j_*}-\alpha_*\right|^2 + \grad_{j_*}\Loss\left(\paramHatkm\right)\left(\paramHatk_{j_*}-\alpha_*\right)\\ &\qquad\qquad\qquad+ \frac{\LipshitzLower}{2}\left\|\Delta^{(k)}\right\|_2^2 +\frac{\LipshitzLower}{2}\left|\paramHatk_{j_*}\right|^2.
\end{aligned}
\nonumber
\end{equation}
This is a contradiction provided that $\frac{\LipshitzLower}{2}\left|\paramHatk_{j_*}\right|^2 + \grad_{j_*}\Loss\left(\paramHatkm\right)\left(\paramHatk_{j_*}-\alpha_*\right)\geq 0$. Later, we will show that $\sgn\left(\grad_{j_*}\Loss\left(\paramHatkm\right)\right)=-\sgn\left(\alpha_*\right)$ and $\LipshitzLower|\alpha_*|\leq\left|\grad_{j_*}\Loss\left(\paramHatkm\right)\right| \leq\LipshitzUpper|\alpha_*|$. With these, if $\frac{\paramHatk_{j_*}}{\alpha_*}\leq 1$, we have $\grad_{j_*}\Loss\left(\paramHatkm\right)\left(\paramHatk_{j_*}-\alpha_*\right)\geq 0$ and the claim follows. Otherwise, we have $\left|\paramHatk_{j_*}\right|\geq \left|\paramHatk_{j_*}\right|-\left|\alpha_*\right|=\left|\paramHatk_{j_*}-\alpha_*\right|$ so that $\left|\paramHatk_{j_*}\right|\geq \frac{2\LipshitzUpper}{\LipshitzLower}\left|\alpha_*\right|$ and hence,
\begin{equation}
\begin{aligned}
\frac{\LipshitzLower}{2}\left|\paramHatk_{j_*}\right|^2 + \grad_{j_*}\Loss\left(\paramHatkm\right)\left(\paramHatk_{j_*}-\alpha_*\right)& \geq \frac{\LipshitzLower}{2}\frac{2\LipshitzUpper}{\LipshitzLower} \left|\alpha_*\right|\left|\paramHatk_{j_*}-\alpha_*\right| - \LipshitzUpper\left|\alpha_*\right|\left|\paramHatk_{j_*}-\alpha_*\right|\\
&= 0.
\end{aligned}
\nonumber
\end{equation}

\bigskip
To get the claimed properties of $\grad_{j_*}\Loss\left(\paramHatkm\right)$, note that
\begin{equation}
\begin{aligned}
\frac{\LipshitzLower}{2}\left|\alpha_*\right|^2 &\leq \Loss\left(\paramHatkm\right) - \Loss\left(\paramHatkm+\alpha_*e_{j_*}\right)\\
&\leq -\frac{\LipshitzLower}{2}\left|\alpha_*\right|^2 -\grad_{j_*}\Loss\left(\paramHatkm\right) \alpha_*\,,
\end{aligned}
\nonumber
\end{equation}
and hence $\sgn\left(\grad_{j_*}\Loss\left(\paramHatkm\right)\right)=-\sgn\left(\alpha_*\right)$ and $\LipshitzLower|\alpha_*|\leq\left|\grad_{j_*}\Loss\left(\paramHatkm\right)\right|$. Also, we can establish
\begin{equation}
\begin{aligned}
\frac{\LipshitzUpper}{2}\left|\alpha_*\right|^2 &\geq \Loss\left(\paramHatkm\right) - \Loss\left(\paramHatkm+\alpha_*e_{j_*}\right)\\
&\geq -\frac{\LipshitzUpper}{2}\left|\alpha_*\right|^2 -\grad_{j_*}\Loss\left(\paramHatkm\right) \alpha_*\,.
\end{aligned}
\nonumber
\end{equation}
Since $-\grad_{j_*}\Loss\left(\paramHatkm\right) \alpha_*\geq 0$, we can conclude that $\left|\grad_{j_*}\Loss\left(\paramHatkm\right)\right|\leq \LipshitzUpper|\alpha_*|$.

This concludes the proof of the lemma.\\
\end{proof}

\begin{lemma}[General Forward Step]
The first time the algorithm reaches a support size of $k$ at the beginning of the forward step, we have
\begin{equation}
\left|\Loss\left(\paramStar\right) - \Loss\left(\paramHatkm\right)\right| \leq \sqrt{2 \left|\suppStar - \suppHatkm\right|\, \LipshitzUpper\, \delta_f^{(k)}}\,\left\|\paramStar - \paramHatkm\right\|_2.
\nonumber
\end{equation}
\label{lem:forwardstep}
\end{lemma}

\begin{proof}
Under the assumption of the lemma, we have
\begin{equation} \nonumber
\Loss\left(\paramHatkm\right) - \inf_{j\notin\suppHatkm,\alpha\in\real}\Loss\left(\paramHatkm + \alpha e_j\right) = \delta_f^{(k)}.
\end{equation}
For any $\eta\in\real$, we have 
\begin{equation}
\begin{aligned}
-\left|\suppStar-\suppHatkm\right|\delta_f^{(k)} &\leq \sum_{j\in\suppStar-\suppHatkm}\Loss\left(\paramHatkm + \eta\paramStar_{j} e_{j}\right) - \Loss\left(\paramHatkm\right)\\
&\leq \eta\sum_{j\in\suppStar-\suppHatkm}\grad_{j}\Loss\left(\paramHatkm\right) \paramStar_{j} + \eta^2\frac{\LipshitzUpper}{2}\left\|\paramStar-\paramHatkm\right\|_2^2\\
&\leq \eta\left(\Loss\left(\paramStar\right)-\Loss\left(\paramHatkm\right)\right) + \eta^2\frac{\LipshitzUpper}{2}\left\|\paramStar-\paramHatkm\right\|_2^2.
\end{aligned}
\nonumber
\end{equation}

Optimizing the RHS over $\eta$, we obtain
\begin{equation}
\begin{aligned}
|\suppStar - \suppHatkm|\delta_f^{(k)} &\geq \frac{\left(\Loss\left(\paramStar\right) - \Loss\left(\paramHatkm\right)\right)^2}{2\LipshitzUpper\,\|\paramStar - \paramHatkm\|_2^2}.
\end{aligned}
\nonumber
\end{equation}

This concludes the proof of the lemma.\\
\end{proof}

\begin{lemma}[General Error Bound]
The first time the algorithm reaches a support size of $k$ at the beginning of the forward step, assuming $RSC\left(|\suppHatk\cup\suppStar|\right)$ holds, we have
\begin{equation}
\left\|\paramHatkm_{\suppHatkm - \suppStar}\right\|_2^2 \le \frac{4\LipshitzUpper|\suppStar \cup \suppHatkm|\delta_f^{(k)}}{\LipshitzLower^2} \left(\frac{\noiseLevel}{\sqrt{\LipshitzUpper\EpsilonStop}}+ \sqrt{\frac{2|\suppStar - \suppHatkm|}{|\suppStar \cup \suppHatkm|}}\,\right)^2\!\!\!\!.
\nonumber
\end{equation}
\label{lem:errorbound}	
\end{lemma}

\begin{proof}
Let 
\begin{equation}
G\left(\delpar\right) \defn \Loss(\paramStar + \delpar) - \Loss(\paramStar) - \sqrt{2|\suppStar - \suppHatkm| \, \LipshitzUpper\,\delta_f^{(k)}} \|\delpar\|_2.
\nonumber
\end{equation}
It can be seen that $G(0) = 0$, and from Lemma~\ref{lem:forwardstep}, $G(\paramHatkm - \paramStar) \le 0$. Further, $G(\delpar)$ is sub-homogeneous (over a limited range): $G(t \delpar) \le t G(\delpar)$ for $t \in [0,1]$. Thus, for a carefully chosen $r > 0$, if we show that $G(\delpar) > 0$ for all $\delpar \in \{\delpar : \|\delpar\|_{2} \le r,\;\|\delpar\|_0 \le |\suppUnion|\}$, where $\suppUnion = |\suppHatk \union \suppStar|$, then it follows that $\|\paramHatk - \paramStar\|_2 \le r$.
If not, then there would exist some $t \in [0,1)$ such that $\|t (\paramHatk - \paramStar)\|_2 = r$, whence we would arrive at the contradiction
\begin{align*}
	0 < G\left(t (\paramHatk - \paramStar)\right) \le t G\left(\paramHatk - \paramStar\right) \le 0.
\end{align*}

Thus, it remains to show that $G(\delpar) > 0$ for all $\delpar \in\{\delpar: \|\delpar\|_{2} \le r,\;\|\delpar\|_0 \le |\suppUnion|\}$. By RSC, we have
\begin{align*}
	\Loss(\paramStar + \delpar) - \Loss(\paramStar) \ge \grad \Loss(\paramStar) \cdot \delpar + \frac{\LipshitzLower}{2} \|\delpar\|_2^2.
\end{align*}
We can establish 
\begin{align*}
	\grad \Loss(\paramStar) \cdot \delpar &\ge  - |\grad \Loss(\paramStar) \cdot \delpar|\\
											&\ge - \|\grad \Loss(\paramStar)\|_\infty \|\delpar\|_1 = -\noiseLevel \|\delpar\|_1,
\end{align*}
and hence,
\begin{align*}
G(\paramStar + \delpar) &\ge - \noiseLevel \|\delpar\|_1 + \frac{\LipshitzLower}{2} \|\delpar\|_2^2 - \sqrt{2 |\suppStar - \suppHatkm| \LipshitzUpper \delta_f^{(k)}} \|\delpar\|_2\\
						&\ge \|\delpar\|_2 \left(  \frac{\LipshitzLower}{2} \|\delpar\|_2 - \noiseLevel \sqrt{|\suppStar \cup \suppHatk|} - \sqrt{2 |\suppStar - \suppHatkm| \LipshitzUpper \delta_f^{(k)}}\right)\\
						&> 0,
\end{align*}
if $\|\delpar\|_2 = r$ for $$r = \frac{2}{\LipshitzLower} \left(\noiseLevel \sqrt{|\suppStar \cup \suppHatk|} + \sqrt{2 |\suppStar - \suppHatkm| \LipshitzUpper \delta_f^{(k)}}\right).$$
Hence,
\begin{equation}
\left\|\paramHatkm_{\suppHatkm - \suppStar}\right\|_2^2 \le \frac{4\LipshitzUpper|\suppStar \cup \suppHatkm|\delta_f^{(k)}}{\LipshitzLower^2} \left(\frac{\noiseLevel}{\sqrt{\LipshitzUpper\delta_f^{(k)}}}  + \sqrt{\frac{2|\suppStar - \suppHatkm|}{|\suppStar \cup \suppHatkm|}}\right)^2.
\nonumber
\end{equation}
Finally, consider the fact that $\delta_f^{(k)}\ge \EpsilonStop$. This concludes the proof of the lemma.
\end{proof}

\end{document}